\documentclass{article}
\usepackage{spconf,amsmath,graphicx}
\usepackage{booktabs}
\usepackage{amssymb}

\title{Role of audio in audio-visual video summarization}
\name{Ibrahim Shoer, Berkay Köprü, Engin Erzin} %\thanks{Thanks to XYZ agency for funding.}}
\address{KUIS-AI Laboratory \\
  Multimedia, Vision and Graphics Group \\
  College of Engineering, Ko\c{c} University, Istanbul, Turkey\\
  {\it ishoer20,bkopru17,eerzin@ku.edu.tr}}
\begin{document}
\ninept
\maketitle
\begin{abstract}
Video summarization attracts attention for efficient video representation, retrieval, and browsing to ease volume and traffic surge problems. Although video summarization mostly uses the visual channel for compaction, the benefits of audio-visual modeling appeared in recent literature. The information coming from the audio channel can be a result of audio-visual correlation in the video content. In this study, we propose a new audio-visual video summarization framework integrating four ways of audio-visual information fusion with GRU-based and attention-based networks. Furthermore, we investigate a new explainability methodology using audio-visual canonical correlation analysis (CCA) to better understand and explain the role of audio in the video summarization task. Experimental evaluations on the TVSum dataset attain F1 score and Kendall-tau score improvements for the audio-visual video summarization. Furthermore, splitting video content on TVSum and COGNIMUSE datasets based on audio-visual CCA as positively and negatively correlated videos yields a strong performance improvement over the positively correlated videos for audio-only and audio-visual video summarization.
\end{abstract}

\section{Introduction}
\label{sec:intro}

Video sharing and streaming services have taken over the traditional media medium, as they provide vast, personalized, and on-demand content. With the success of these platforms, internet usage in terms of both upload and download inflated. Recently, the video summarization field attracted researchers as it produces solutions for efficient video representation, retrieval, and browsing to ease the complications caused by video content and traffic surge. 

Traditionally video summarization processes visual information and generates a meaningful subset of the original video, which covers the most salient segments \cite{vssurvey, vssurvey2}. Video summarization studies can be categorized into optimization-based weakly supervised or unsupervised and supervised studies. Optimization-based video summarization studies formulate video summary generation via optimization of hand-crafted heuristics such as representativeness, diverseness, etc., which are derived from the properties of the frame-level visual information \cite{earlyvs_lu2013story, earlyvs_kim2018, cizmeciler2022queryvs}. An early study selects sub-shots that maximize the weighted combination of self-defined story, importance, and diversity objectives \cite{earlyvs_lu2013story}. In a later study, video summarization is formulated as a sequential decision-making process, and it is solved using deep reinforcement learning with self-defined representativeness and diversity rewards  \cite{earlyvs_kim2018}. In a more recent study \cite{cizmeciler2022queryvs}, similar to \cite{earlyvs_kim2018}, representativeness and diverseness are considered, and a relevance term is introduced to create query-specific summaries using semantic saliency maps.

With the availability of annotated datasets such as TVSum \cite{tvSum2018} and COGNIMUSE \cite{Cognimuse2017}, recent studies focus on developing supervised solutions which formulate video summarization as a sequence-to-sequence learning task where the input sequences are frame-level visual representations \cite{kopru2021affectivevs,ji2019attvs, zhao2021audiovisualvs}. In \cite{kopru2021affectivevs}, a two-step approach is adopted into summarization of human-centric videos. Affective information represented as embeddings extracted from a trained emotion recognition is injected using attention mechanisms into an encoder-decoder-like fully convolutional neural network, which is trained using supervised learning. In \cite{ji2019attvs}, video summarization formulated as a sequence-to-sequence translation and proposed an attentive encoder-decoder based on long-short term memory (LSTM) networks and attention mechanism.

Recently multi-modal approaches for video summarization have attracted significant attention \cite{zhao2021audiovisualvs, ZHAO2022360audiovisualvs}. In \cite{zhao2021audiovisualvs}, embeddings from a pre-trained VGGISH network represent audio information, and embeddings from pre-trained GoogLeNet represent visual information. First, these two separate information flows are fed to parallel LSTMs, then the outputs of these LSTMs are fused and passed to a final LSTM generating the importance scores of the frames. In \cite{ZHAO2022360audiovisualvs}, the approach in \cite{zhao2021audiovisualvs} is followed to represent the audio-visual information, and LSTMs are replaced with transformers. Both of these studies present improvements in video summarization using audio-visual representations over the SumMe and TVSum datasets, which are both annotated over the visual channel without audio modality. This raises the question of when and how audio modality contributes to the video summarization task. In this study, we investigate these questions by considering a self-supervised audio representation that integrates with a visual representation for audio-visual video summarization (AV-VSUM).

 In the proposed AV-SUM, a window of audio is represented with the embeddings extracted from Wav2Vec 2.0 \cite{baevski2020wav2vec}, and in parallel visual information is represented with the embeddings from GoogLeNet \cite{googlenet}. Then we pre-train fully-convolutional encoder-decoder architectures for each modality for the summarization task. Using these pre-trained unimodal networks, we construct several late fusion schemes that integrate two-stream of information from either the attention mechanism or the Gated-Recurrent Units (GRU) layers. To summarize the main contributions of this study:\begin{enumerate}
\item We propose a new self-supervised audio representation followed by a convolutional feature extractor.
\item We propose a novel pre-training strategy that improves multi-modal video summarization performance on all evaluation metrics.
\item We investigate a new explainability methodology using audio-visual canonical correlation analysis to understand when and how audio modality contributes to the video summarization task.
\end{enumerate}

%Recently multi-modal approaches for video summarization have attracted significant attention \cite{zhao2021audiovisualvs, ZHAO2022360audiovisualvs}. % and \cite{xie2022multimodalvs} In \cite{zhao2021audiovisualvs}, embeddings from a pre-trained VGGISH network represent aural information, and embeddings from pre-trained GoogLeNet represent visual information. First, these two separate information flows are fed to parallel LSTMs, then the output of these LSTMs is fused and passed to a final LSTM generating the importance scores of the frames. In \cite{ZHAO2022360audiovisualvs}, the approach in \cite{zhao2021audiovisualvs} is followed to represent the aural and visual information, and LSTMs are replaced with Transformers. However, none of these studies consider auditory representation based on self-supervised learning, which is shown in \cite{} to be beneficial for multiple tasks. In addition, while neither of \cite{zhao2021audiovisualvs} and \cite{ZHAO2022360audiovisualvs}, generate performances on a dataset designed for multi-modal video summarization like COGNIMUSE, their performances are not cross-validated as in \cite{kopru2021affectivevs} showed the performances of the solutions are decreased when it is cross-validated. % In \cite{xie2022multimodalvs}, audio, visual, and text modalities are adopted for the video summarization. 

The rest of this paper organized as follows. Section 2 describes main building blocks of the proposed framework. Section 3 presents the experiments conducted together with the performance evaluations. Finally, conclusion is presented in Section 4.

\section{Methodology}

\maketitle
\maketitle
\label{sec:method}
% \subsection{Approach}
Video summarization addresses the problem of selecting salient video segments or key frames. Although ground truth summarizations are typically constructed by human annotations over the visual frames, the audio channel carries useful information for the summarization task, as demonstrated in recent literature \cite{zhao2021audiovisualvs}. The information coming from the audio channel can be expected as a result of audio-visual correlation in the video content. In this study, we define a new audio-visual video summarization framework integrating four ways of audio-visual information fusion and later evaluate performances of these frameworks to better understand and explain the role of audio channel for the summarization task.

In the following, we state the summarization problem, define audio-visual representations, and present the unimodal and multimodal video summarization frameworks.

\subsection{Problem Statement}
In video summarization, key frame selection is often performed in two schemes, either by assigning binary labels or by assigning frame-level importance scores \cite{rochan2018video}. In this paper, we follow the binary label assignment formulation. The video summarization first selects $N$ number of frames after a uniform decimation in time and then assigns binary labels to these frames. 
Hence the summarization network receives an input feature matrix, $\mathbf{f} \in \mathbb{R}^{N \times D}$, and emits an output matrix, $\mathbf{s} \in \mathbb{R}^{N \times 2}$, where $D$ is the dimensionality of the frame-level visual feature. The output $\mathbf{s}$ with 2-dimensional nodes represents the positive and negative classes for the key-frame selection, and these two nodes output class probability values at the output of the network. Then the positive class for key-frame selection or the negative class for frameskip is set by picking the node with a higher probability. Eventually, the video summary is constructed from the key-frames that are labeled as positive.

\subsection{Feature Extraction}
Visual features are defined as the pool5 layer embeddings of the pre-trained \emph{GoogleNet} for each visual frame and represented as $\mathbf{f}_{v}^n \in \mathbb{R}^{D}$ at frame $n$ with dimension $D = 1024$.

We consider a three-second temporal acoustic window centered at each frame to extract the audio features. The raw audio embeddings are extracted from block 15 of the pre-trained \emph{wav2vec2.0 Large} model and represented as $\mathbf{f}_{w} \in \mathbb{R}^{N \times M \times D}$ , where $M = 149$ is the number of acoustic features from the three-second window, and feature dimension is same as the visual features, $D = 1024$. Block 15 embeddings are preferred since they deliver the lowest phone error rate on \emph{Librispeech} \cite{baevski2021unsupervised}.

\begin{figure*}[h]
\centering
\includegraphics[width=\textwidth]{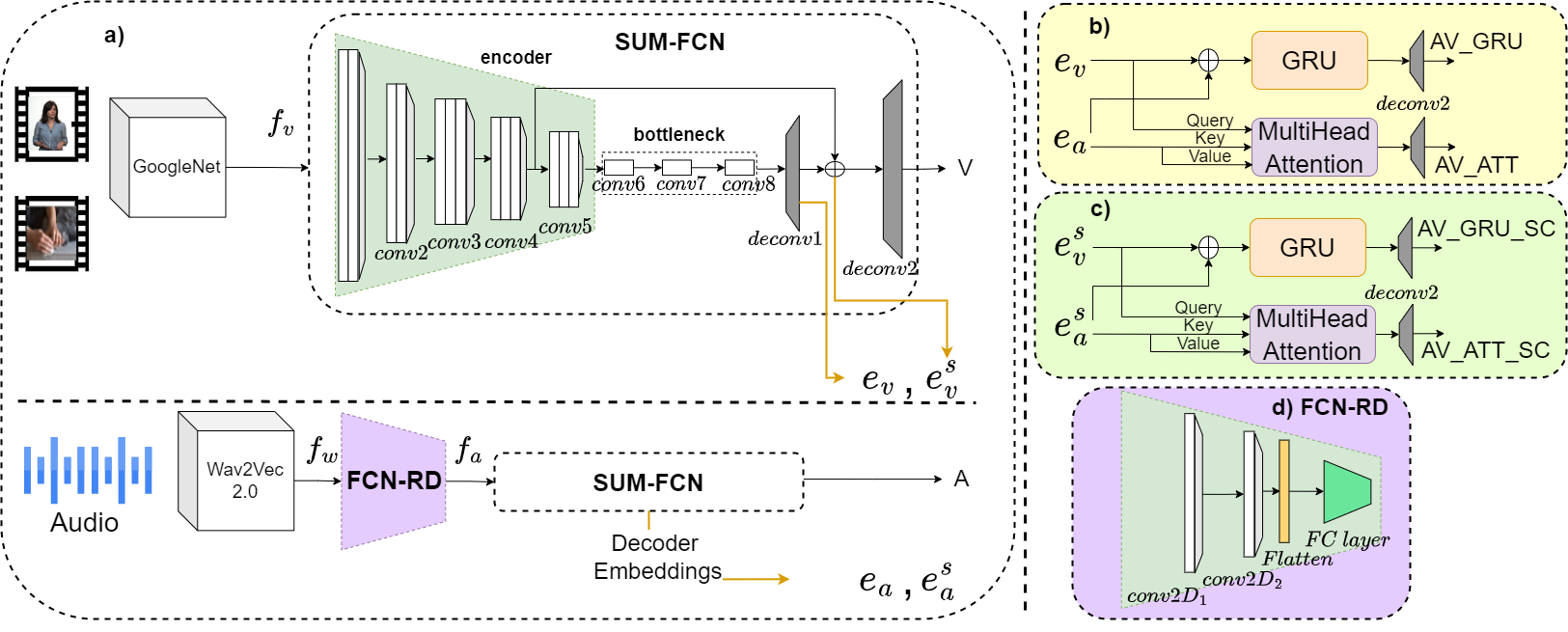}
\caption{Block diagram of the proposed audio-visual video summarization model: (a) Unimodal audio and visual SUM-FCN networks, (b) late fusion of embeddings before the skip connection, (c) late fusion of embeddings after the skip connection, (d) FCN-RD - convolutional audio feature extraction network}
\label{fig:diagram}
\end{figure*}

The raw acoustic feature $\mathbf{f}_w^n$ at frame $n$ is then fed to a convolutional network (FCN-RD) to obtain audio features $\mathbf{f}_a^n \in \mathbb{R}^{D}$. As depicted in Figure~\ref{fig:diagram}(d), the FCN-RD network consists of two convolutional blocks; each block is formed of a 2D convolution layer and \emph{maxpool} layer. After the convolutional blocks, there is a flatten layer followed by a fully connected layer at the end.

% \{$h_{i}$, $h_{i+1}$, $h_{i+2}$, …,$h_{N}$\} 
% \subsection{Feature Extraction}

\subsection{Video Summarization}
SUM-FCN is a popular fully convolutional network-based video summarization architecture trained on visual features \cite{rochan2018video}. In the multimodal audio-visual setting, we adapt SUM-FCN for both video and audio modalities. SUM-FCN network trained on visual modality is considered the baseline. Decoder-level embeddings from both models are extracted and fused together in a multimodal setting. In the following, we briefly describe unimodal SUM-FCN for audio and video features and present four possible fusion mechanisms for the audio-visual video summarization problem.

\subsubsection{Unimodal SUM-FCN}
SUM-FCN follows an encoder-decoder architecture. The encoder block consists of fully convolutional layers. A series of frame-level features are fed to the encoder block to obtain high-level representations. Which is then processed by bottleneck layers. On the other hand, the decoder block consists of two deconvolution layers.

Figure~\ref{fig:diagram}(a) presents unimodal SUM-FCN networks for visual and audio channels. The visual SUM-FCN network receives the sequence of visual features $\mathbf{f}_v = [\mathbf{f}_{v}^1, \mathbf{f}_{v}^2,....,\mathbf{f}_{v}^N] \in \mathbb{R}^{N \times D}$ and outputs a visual summarization sequence $\mathbf{s}_v \in \mathbb{R}^{N \times 2}$.

The audio SUM-FCN network similarly receives the sequence of audio features $\mathbf{f}_a \in \mathbb{R}^{N \times D}$ and outputs an audio summarization sequence $\mathbf{s}_a \in \mathbb{R}^{N \times 2}$.
 
\subsubsection{Multimodal Fusion Networks}

The audio-visual video summarization network is constructed as a late-fusion of unimodal audio and visual SUM-FCN networks.
The decoder block in SUM-FCN consists of two deconvolution layers, with a skip connection in between. We consider using two sets of embeddings obtained after the first deconvolution layer for the late fusion. The first set of embeddings, $\mathbf{e}_v$ and $\mathbf{e}_a$ respectively for visual and audio channels, is collected before the skip connection as depicted in Figure~\ref{fig:diagram}(a). Whereas the second set of embeddings, $\mathbf{e}_v^s$ and $\mathbf{e}_a^s$, is collected after the skip connection.

We proposed GRU-based and attention-based audio-visual fusion networks. The GRU-based fusion network receives audio-visual embeddings before or after the skip connections and passes the output to the last deconvolution layer to extract video summarization labels. These two fusion networks are denoted as AV\_GRU and AV\_GRU\_SC, respectively, with their embeddings before and after the skip connections.

The attention-based fusion network similarly receives the same embeddings.In the attention-based fusion, video embeddings $\mathbf{e}_v$ and $\mathbf{e}_v^s$ as query and audio embeddings $\mathbf{e}_a$ and $\mathbf{e}_a^s$ as key and value. Query, key, and value are given as input to a Multi-head attention layer, and then pass the output to the last deconvolution layer to extract video summarization labels. Attention-based fusion networks are denoted as AV\_ATT and AV\_ATT\_SC, respectively, with their embeddings before and after the skip connections.

Figure~\ref{fig:diagram}(b)-(c) depict the proposed GRU-based and attention-based late-fusion networks.

\section{Experimental Evaluations}
%\maketitle
%\maketitle
\label{sec:results}

The proposed audio-visual video summarization architecture in Section 2 is evaluated on the TVSum and COGNIMUSE datasets. In this section, first, datasets and implementation details are introduced. Then the proposed CCA-based  audio-visual correlation analysis is described. Finally, the performances of the unimodal and multimodal architectures are provided with discussions.

\subsection{Datasets}

We train and evaluate video summarization models on the widely used TVSum \cite{tvSum2018} and COGNIMUSE \cite{Cognimuse2017} datasets. The TVSum dataset is collected from YouTube, covering 10 different categories, such as making sandwiches, vehicle tires, etc., including in total of 50 videos. The annotations are in terms of both 5-level importance scores and binary labels, where each shot is rated by 20 annotators. The COGNIMUSE dataset contains the last half-hours of 7~Hollywood movies. The video summary annotations are binary and rated by 3 annotators.

\subsection{Models Training}

The visual SUM-FCN is trained to map the visual features $\mathbf{f}_v$ to the summarization label sequence  $\mathbf{s}_v$ by optimizing cross-entropy loss with the ground truth labels. Similarly, the audio SUM-FCN is trained to map $\mathbf{f}_a$ to $\mathbf{s}_a$ by optimizing cross-entropy loss.
For multimodal fusion models, we freeze all layers before the extraction of embeddings and only fine-tune the newly added layers and the last deconvolution layer. All models are trained using Adam optimizer, with a learning rate of $1e-3$ for unimodal models and $1e-4$ for multimodal fusion.

Considering the imbalanced nature, all the models are trained using a weighted binary cross-entropy loss which is defined as 
\begin{equation}
    L_{\text{SUM}} = -\frac{1}{N}\sum^{N}_{j=1} w_{z_{j}}(z_{j}\log{s_j[0]}+(1-z_{j})\log{(s_j[1]))},
\end{equation}
where $z_j \in {0,1}$ is the binary ground truth, $s$ is the predicted score and $w_{z_{j}}$ is the weight of the $j^{\text{th}}$ frame. The weights for the binary target are defined as
\begin{equation}
    w_{0} = \frac{1}{N}\sum_{j=1}^N z_j \;\;\;  \mathrm{and} \;\;\;
    w_{1} = 1 - w_{0}.
\end{equation}

We performed cross-validation for both datasets. For TVSum, we have five folds, with one-fifth of the video in the test set and the rest for training. For COGNIMUSE, we perform leave-one-out cross-validation.

\subsection{Canonical Correlation Analysis}

Although summarization annotations are typically performed by human annotators over the visual frames, the audio channel is observed to carry useful information for the summarization task. We hypothesize finding the source of this useful information from audio carried for summarization by using audio-visual canonical correlation analysis (CCA).

Hence, we first perform CCA across audio-visual representations, which are defined as $\mathbf{f}_a$ and $\mathbf{f}_v$, respectively. The audio-visual feature space is mapped to a canonical space with transformations maximizing the cross-correlation in the canonical space. Then, we compute audio-visual correlations for each video in the canonical space and sort the videos with respect to these correlation scores. Then the video dataset is split into two sets, one with positively correlated videos (denoted as CCA+) and the other with negatively correlated videos (denoted as CCA-). Performance evaluation of video summarization models has been carried out over these two video subsets, as well as the full video set so that a common performance behavior can be extracted to explain the role of audio in the video summarization task.

For the TVSum dataset, CCA+ set contains 21 videos, and CCA- set contains 24 videos. For the COGNIMUSE dataset, CCA- set contains 4 videos, and CCA+ set contains 3 videos.

\subsection{Results and Discussion}

We evaluate the F1 score performances of the proposed video summarization models across the ground truth and predicted frame-level summarization labels. Table~\ref{tab:1} presents F1 score performances for all videos, for negatively correlated (CCA-) videos, and for positively correlated (CCA+) videos over the TVSum dataset. F1 score performances indicate that the multimodal AV\_GRU model outperforms all other models on all videos and positive correlated (CCA+) video sets. Another important observation is that the GRU-based fusion model outperform the visual SUM-FCN model by more than 2\% for the positive correlated (CCA+) video set. Although videos are annotated only on the visual channel, the audio provides sufficiently rich information for the video summarization task. On the other hand, the visual SUM-FCN model exhibits similar F1 score performances for all three video sets, which suggests the independence of the visual summarization performance from the audio-visual correlation scores. F1 score differences across CCA- and CCA+ video sets reach up to 3.14\% for the audio-based summarization models, which suggests the positively correlated audio-visual channels capture significantly important information from the audio channel for video summarization.
\begin{table}
\centering
\caption{F1 score performances for negatively correlated, positively correlated and all videos on TVSum database}
\begin{tabular}{lccc}

\toprule
Model &       All (\%) &   CCA- (\%) &   CCA+ (\%) \\
\midrule
V           & 57.17     & {\bf 57.55}   & 56.73 \\
A           & 56.90     & 56.11         & 57.80 \\
AV\_GRU     & {\bf 57.66} & 56.19       & {\bf 59.33} \\
AV\_GRU\_SC & 56.76     & 56.21         & 57.39 \\
AV\_ATT     & 56.25     & 55.97         & 56.58 \\
AV\_ATT\_SC & 55.97     & 55.57         & 56.42 \\

\bottomrule
\end{tabular}
\label{tab:1}
\end{table}
%V          &  0.571700 &  0.575528 &  0.567325 \\
% A2D        &  0.569026 &  0.561148 &  0.578030 \\
% Att1\_A2D\_V &  0.562515 &  0.559650 &  0.565788 \\
% Att2\_A2D\_V &  0.559671 &  0.555689 &  0.564222 \\
% GRU1\_A2D\_V &  0.576593 &  0.561945 &  0.593334 \\
% GRU2\_A2D\_V &  0.567596 &  0.562109 &  0.573866 \\

Table~\ref{tab:2} presents the Kendall rank correlation coefficient (Kendall's $\tau$) for TVSum dataset. Interestingly, unimodal audio SUM-FCN and multimodal fusion models outperform the unimodal visual SUM-FCN. The audio SUM-FCN and the GRU-based fusion with skip connection model AV\_GRU\_SC attain a large improvement on Kendall's $\tau$ over the visual SUM-FCN model. A noticeable improvement for Kendall's $\tau$ is observed for the CCA+ video set compared to the CCA- video set, except for V and AV\_ATT models. Results suggest that audio modality is introducing value-added information to improve the rank order performance of the video summarization task.
\begin{table}
\centering
\caption{Kendall's $\tau$ correlation performances for negatively correlated, positively correlated and all videos on TVSum database}
\begin{tabular}{lccc}
\toprule
Model &       All  &   CCA-  &   CCA+ \\
\midrule
V           &  0.009996 &  0.011396 &  0.008397 \\
A          &  0.019713 &  {\bf 0.018902} &  0.020639 \\
AV\_GRU     &  0.014001 &  0.011243 &  0.017153 \\
AV\_GRU\_SC &  {\bf 0.020419} &  0.017914 &  {\bf 0.023282} \\
AV\_ATT     &  0.011144 &  0.011502 &  0.010736 \\
AV\_ATT\_SC &  0.017160 &  0.017136 &  0.017189 \\
\bottomrule
\end{tabular}
\label{tab:2}
\end{table}
% V          &  0.009996 &  0.011396 &  0.008397 \\
% A2D        &  0.019713 &  0.018902 &  0.020639 \\
% Att1\_A2D\_V &  0.011144 &  0.011502 &  0.010736 \\
% Att2\_A2D\_V &  0.017160 &  0.017136 &  0.017189 \\
% GRU1\_A2D\_V &  0.014001 &  0.011243 &  0.017153 \\
% GRU2\_A2D\_V &  0.020419 &  0.017914 &  0.023282 \\

Table~\ref{tab:3} presents F1 score performances over the COGNIMUSE dataset for the three video sets, which are all, CCA-, and CCA+. In the COGNIMUSE dataset, the visual SUM-FCN model outperforms all the other models for the three video sets. Yet the F1 score in CCA+ video set is higher than the F1 score in CCA- video set for all unimodal and multimodal summarization models. This again suggests that positively correlated audio-visual content has the advantage of sustaining better video summarization performance.
\begin{table}
\centering
\caption{F1 score performances for negatively correlated, positively correlated and all videos on COGNIMUSE database}

\begin{tabular}{lccc}
\toprule
Model &       All (\%) &   CCA- (\%) &   CCA+ (\%) \\
\midrule
V           &  {\bf 47.13} &  {\bf 45.63} &  {\bf 49.13} \\
A           &  45.51 &  43.41 &  48.31 \\
AV\_GRU     &  44.51 &  42.03 &  47.82 \\
AV\_GRU\_SC &  44.30 &  42.06 &  47.28 \\
AV\_ATT     &  44.02 &  41.93 &  46.81 \\
AV\_ATT\_SC &  43.94 &  42.00 &  46.52 \\
\bottomrule
\title{Cognimuse}
\end{tabular}
\label{tab:3}
\end{table}

\section{Conclusion}

%\maketitle
%\maketitle
\label{sec:conclusion}

In this study, we investigate the effect of audio on the multimodal video summarization task. First, we train unimodal video summarization architectures utilizing visual and audio modalities. Then, we propose fusion schemes based on attention and GRU combining the representations of the unimodal architectures. To analyze the effect of the audio, we define a CCA-based audio-visual correlation analysis that categorizes videos into positively and negatively correlated classes. GRU-based fusion schemes outperform attention-based schemes by at least 0.5\% and 0.3\% in the F1 score, respectively, on the TVSum and COGNIMUSE datasets. This suggests that with the limited available data and longer time steps, GRU is more successful at learning temporal relations. In addition, our results depict that all the architectures perform significantly better on the positively correlated videos. Hence, salient regions in the positively correlated videos are observed to be more separable than non-salient regions and, in fact, easier to learn. The observed insight on positively correlated audio-visual content may lead to valuable future studies to better fine-tune audio-visual video summarization.
\bibliographystyle{IEEEbib}
\bibliography{ref}

\end{document}